\documentclass[11pt]{article}

\usepackage[preprint]{acl}
\usepackage{times}
\usepackage{latexsym}
\usepackage{booktabs}
\usepackage[T1]{fontenc}
\usepackage[utf8]{inputenc}
\usepackage{microtype}
\usepackage{inconsolata}
\usepackage{graphicx}

\usepackage{microtype}
\usepackage{graphicx}
\usepackage{subfigure}
\usepackage{booktabs} % for professional tables

\usepackage{xcolor}
\usepackage{xspace}
\usepackage{tcolorbox}

\definecolor{lightBlue}{rgb}{0.78, 0.85, 1.0}
\definecolor{lightRed}{rgb}{1.0, 0.85, 0.85}
\newtcbox{\bluebox}{on line, box align=base, colback=lightBlue,colframe=white,size=fbox,arc=3pt, before upper=\strut, top=-2pt, bottom=-4pt, left=-2pt, right=-2pt, boxrule=0pt}
\newtcbox{\redbox}{on line, box align=base, colback=lightRed,colframe=white,size=fbox,arc=3pt, before upper=\strut, top=-2pt, bottom=-4pt, left=-2pt, right=-2pt, boxrule=0pt}

\usepackage{hyperref}
\usepackage{multicol}
\usepackage{multirow}

\usepackage{amsmath}
\usepackage{amssymb}
\usepackage{mathtools}
\usepackage{amsthm}

\usepackage[capitalize,noabbrev]{cleveref}

\theoremstyle{plain}

\theoremstyle{definition}

\theoremstyle{remark}

\usepackage{colortbl}
\usepackage{pgf} % for calculating the values for gradient
\usepackage{xcolor} % enables the use of cellcolor
\usepackage{graphicx}
\usepackage{booktabs} % for better table formatting (optional)

\definecolor{lightOliveGreen}{rgb}{0.75, 0.85, 0.65} % Light Olive Green (adjust as necessary)
\definecolor{lightTomatoRed}{rgb}{1.0, 0.7, 0.6}     % Light Tomato Red (adjust as necessary)

\newtcbox{\greenbox}{on line, box align=base, colback=lightOliveGreen, colframe=white, size=fbox, arc=3pt, before upper=\strut, top=-2pt, bottom=-4pt, left=-2pt, right=-2pt, boxrule=0pt}
\newtcbox{\tomatobox}{on line, box align=base, colback=lightTomatoRed, colframe=white, size=fbox, arc=3pt, before upper=\strut, top=-2pt, bottom=-4pt, left=-2pt, right=-2pt, boxrule=0pt}

\definecolor{high}{HTML}{6B8E23}  % The color for increases (Olive Green)
\definecolor{lowgreen}{HTML}{F0F8E7}  % A very light green for the lower end of the gradient
\definecolor{redhigh}{HTML}{FF6347} % The color for decreases (Red, Tomato)
\definecolor{lowred}{HTML}{FFE6E6}  % A very light red for the lower end of the gradient
\newcommand{\opacity}{50}         % Opacity set to 50 (higher opacity, but you can reduce further if needed)

\newcommand*{\minvaldiffa}{0}% define the minimum value on your data set
\newcommand*{\maxvaldiffa}{50}% define the maximum value in your data set!
\newcommand{\gcb}[1]{
    \pgfmathparse{min(#1,\maxvaldiffa)}%
    \xdef\clipval{\pgfmathresult}%
    \pgfmathparse{max(\clipval,\minvaldiffa)}%
    \xdef\clipval{\pgfmathresult}%
    \pgfmathparse{int(round(100*(\clipval/(\maxvaldiffa-\minvaldiffa))-(\minvaldiffa*(100/(\maxvaldiffa-\minvaldiffa)))))}%
    \xdef\tempa{\pgfmathresult}%
    \cellcolor{high!\tempa!lowgreen!\opacity}%
    $#1$%
}

\newcommand*{\minvaldiffb}{0}% define the minimum value on your data set
\newcommand*{\maxvaldiffb}{50}% define the maximum value in your data set!
\newcommand{\gca}[1]{
    \pgfmathparse{min(#1,\maxvaldiffb)}%
    \xdef\clipval{\pgfmathresult}%
    \pgfmathparse{max(\clipval,\minvaldiffb)}%
    \xdef\clipval{\pgfmathresult}%
    \pgfmathparse{int(round(100*(\clipval/(\maxvaldiffb-\minvaldiffb))-(\minvaldiffb*(100/(\maxvaldiffb-\minvaldiffb)))))}%
    \xdef\tempa{\pgfmathresult}%
    \cellcolor{redhigh!\tempa!lowred!\opacity}%
    $#1$%
}

\title{In-Context Learning May Not Elicit Trustworthy Reasoning: A-Not-B Errors in Pretrained Language Models}

\author{%
    Pengrui Han\thanks{Equal contribution.}\thanks{Work done as intern.}$^{\,1,3}$, Peiyang Song\footnotemark[1]$^{\,2}$, Haofei Yu$^{\,3}$, Jiaxuan You$^{\,3}$ \\
    $^1$Carleton College, ~$^2$California Institute of Technology\\
    $^3$University of Illinois Urbana-Champaign\\
    \texttt{barryhan@carleton.edu}, \texttt{psong@caltech.edu} \\
}

\newcommand{\llamalarge}{{\texttt{Llama3-70b}}\xspace}
\newcommand{\llamasmall}{\texttt{Llama3-8b}\xspace}
\newcommand{\llamatwo}{\texttt{Llama2-70b}\xspace}
\newcommand{\qwenlarge}{\texttt{Qwen1.5-72b}\xspace}
\newcommand{\qwensmall}{\texttt{Qwen1.5-7b}\xspace}

\newcommand{\mathqa}{\texttt{MathQA}\xspace}
\newcommand{\commonsenseqa}{\texttt{CommonsenseQA}\xspace}
\newcommand{\winogrande}{\texttt{Winogrande}\xspace}
\newcommand{\sciq}{\texttt{SciQ}\xspace}

\begin{document}
\maketitle

\begin{abstract}
 Recent advancements in artificial intelligence have led to the creation of highly capable large language models (LLMs) that can perform tasks in a human-like manner. However, LLMs exhibit only infant-level cognitive abilities in certain areas. One such area is the A-Not-B error, a phenomenon seen in infants where they repeat a previously rewarded behavior despite well-observed changed conditions. This highlights their lack of inhibitory control -- the ability to stop a habitual or impulsive response. In our work, we design a text-based multi-choice QA scenario similar to the A-Not-B experimental settings to systematically test the inhibitory control abilities of LLMs. We found that state-of-the-art LLMs (like \llamasmall) perform consistently well with in-context learning (ICL) but make errors and show a significant drop of as many as 83.3\% in reasoning tasks when the context changes trivially. This suggests that LLMs only have inhibitory control abilities on par with human infants in this regard, often failing to suppress the previously established response pattern during ICL. 
\end{abstract}

\section{Introduction}
\label{sec:intro}

\begin{figure}[t]
\centering
\includegraphics[width=\columnwidth]{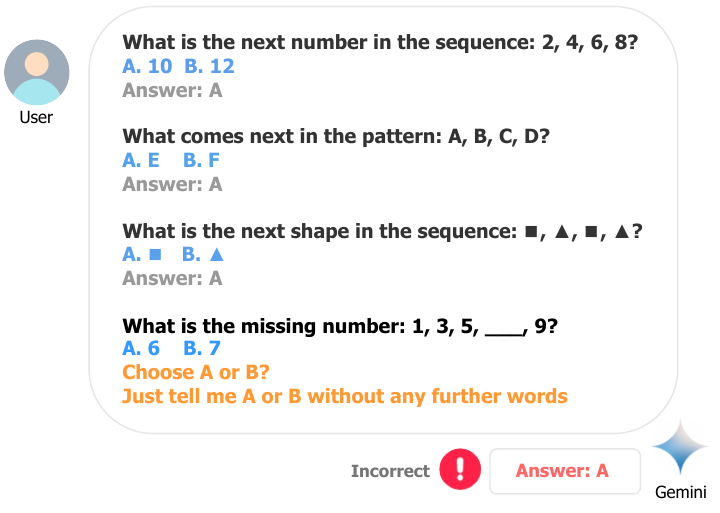}
\caption{\textbf{A-Not-B Style Few-Shot Prompts Mislead \texttt{Gemini} on Elementary School-Level Questions.} This figure presents a simple few-shot prompt that tricks an advanced model \texttt{Gemini} on elementary school-level questions by consistently providing examples with Answer A. This example was collected on \textit{Sep 21, 2024}, with the possibility that future updates may lead to different results. We provide a screenshot in Appendix \S\ref{sec:gemini}.}
\label{fig:AnotB}
\end{figure}

In the field of cognitive science, there is a classic phenomenon called the A-Not-B error~\cite{popick2011investigating, smith2005development, vorms2012not}. In a typical A-Not-B task, an experimenter repeatedly hides an attractive toy at one location (e.g., under box A) within the baby’s reach while the baby watches, and has the baby retrieve it several times. Then, during the critical trial, the experimenter moves the toy to a new location (e.g., under box B) that is also within the baby's reach \textit{while the baby watches}, yet the baby fails to understand the straightforward change of situation, and continues previous actions to search under the original location (box A). This is a key indication of infants' limited \textit{inhibitory control}: \textit{the ability to stop or control impulsive responses and instead use deliberate attention and reasoning when the context changes} ~\cite{geier2013adolescent, fiske2019neural}. On the other hand, adults have developed strong inhibitory control abilities over the years as they have matured~\cite{geier2013adolescent}. For instance, an adult may initially reach for a frequently used coffee mug on its usual shelf but quickly adapt and reach for it on a new shelf after remembering it was moved the day before. Similarly, evidence has shown that when infants grow to about 12 months \cite{anotbtimingrate}, they typically begin to have such inhibitory control abilities, and no longer exhibit A-not-B errors.

Recent advancements in Large Language Models (LLMs) \cite{saravanan2023exploring, zhang2024mmllmsrecentadvancesmultimodal, 10.1145/3641289, KASNECI2023102274, zhao2023surveylargelanguagemodels} have significantly impacted cognitive science-related research \citep{feng2024far}. Researchers find that LLMs perform well in some cognitive tasks, such as Theory-of-mind \citep{kosinski2023theory}, analogical reasoning \citep{webb2023emergent}, and moral reasoning \citep{almeida2024exploring}, which proves that LLMs have some basic cognitive abilities similar with humans. However, they still struggle with other, sometimes embarrassingly simple cognitive settings, including tasks like spatial reasoning \citep{sharma2023exploring, Yang_2019_ICCV}, solving cunningly designed problems \citep{nezhurina2024alice}, and reversal learning \citep{berglund2024reversal}, indicating a lack of robust cognitive abilities. Such failure cases always suggest meaningful directions to improve LLM abilities and alignment with human intelligence. These successes and failures of LLMs in cognitive abilities are better understood through insights from cognition-related theories, including attention \citep{vaswani2017attention}, working memory \citep{li2022large}, long-term memory \citep{zhong2024memorybank}, chain-of-thought prompting \citep{wei2023chainofthoughtpromptingelicitsreasoning} and more. In this work, we draw inspiration from human A-Not-B error experiments to test the inhibitory control ability of LLMs. Our objective is to assess how well LLMs can inhibit established patterns in contexts and to explore strategies for enhancing this ability in practical applications.

To test the inhibitory control ability of LLMs, we adapt the A-Not-B experimental setting to be purely language-based. Like infants developing a habitual response by seeing a toy repeatedly placed in box A, we show LLMs several multiple-choice questions with the same correct option to establish a trivial pattern. Then, during the critical trial, we present a new question where the correct answer is not A (e.g. B). We refer to this prompting construction strategy as \textit{A-Not-B prompting}. Our goal is to see if LLMs can break the established trivial pattern and use deliberate attention and reasoning, similar to humans' inhibitory control. 

Surprisingly, as illustrated in Figure \ref{fig:AnotB}, even advanced LLMs like \texttt{Gemini} \cite{geminiteam2024gemini} fail to answer extremely simple questions correctly under this scenario. In our further experiments, by only using a 3-shot A-Not-B prompt, popular LLMs (eg. \llamasmall and \qwenlarge) show a significant decrease in accuracy by as much as 83.3\% on some reasoning tasks. This indicates that LLMs have weaker cognitive abilities than elementary school children and can cause serious errors when being used in applications. 

Furthermore, we perform an in-depth analysis of why significant errors occur in LLMs with simple A-Not-B prompting. Our findings suggest that during the training stage, model size and training data quality are critical factors—larger models and higher-quality training data can help reduce these errors. In the post-training phase, strategies like self-explanation \cite{journals/corr/abs-2311-06985} have also been effective in mitigating these errors to a certain degree, though not enough to get rid of the errors. Additionally, the type of task and the number of prompt examples significantly impact performance; LLMs are more likely to rely on superficial A-Not-B patterns, particularly in reasoning tasks that involve less straightforward logical patterns and memorization, and when the A-Not-B pattern is more strongly reinforced. Moreover, LLMs are much less resilient to these errors than human adults, indicating that there may be fundamental differences between LLMs' and humans' reasoning processes.

To our knowledge, we are the first to systematically evaluate the inhibitory control abilities of LLMs. We open-source all code and results under a permissive MIT license, to encourage reproduction and further research exploration \footnote{\href{https://github.com/Peiyang-Song/LLM-A-Not-B-Errors}{https://github.com/Peiyang-Song/LLM-A-Not-B-Errors}}.

\begin{figure*}[t]
    \centering
    \includegraphics[width=1.0\textwidth]{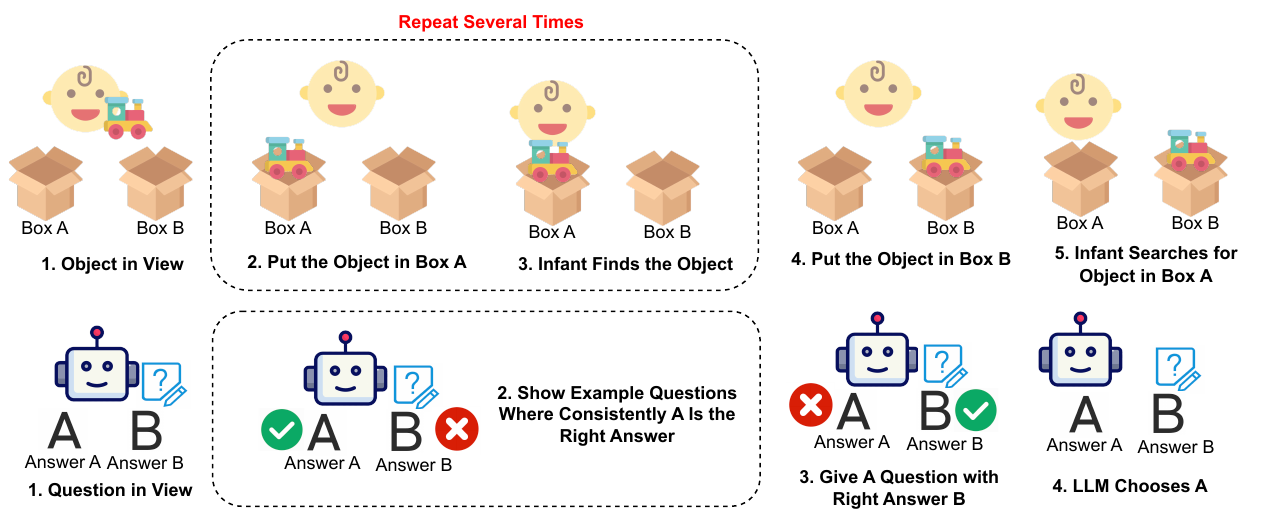}
    \caption{\textbf{Illustration of A-not-B Task Performance in Infants and LLMs.} This figure demonstrates the typical A-not-B error using a binary (Boxes A and B) setup. The first sequence showcases an infant's repeated actions: seeing an object placed in Box A, observing it moved to Box B, yet continuing to search in Box A. This depicts the cognitive phenomenon where prior experience overrides current explicit visual cues. On the bottom sequence, the figure analogously presents a scenario in which LLMs are misled by a consistent answer pattern, illustrating the A-not-B prompting scenarios for LLMs, where they fail to adapt to minimally changed circumstances.}
    \label{fig:AnotB_cog}
\end{figure*}

\section{Related Work}
\label{sec:related}

\paragraph{A-Not-B Error and Inhibitory Control.} The A-Not-B error is a classic cognitive phenomenon observed in infants typically between 8 to 12 months, and resolves as they grow older \cite{popick2011investigating, smith2005development, vorms2012not}. Researchers note this as a significant milestone in cognitive development in humans that signifies the emergence of inhibitory control \cite{diamond2013executive,casey2000structural}. Inhibitory control is the ability to inhibit a response -- such as refraining from seeking an object at a previously habitual location A after explicitly seeing it moved to location B. This ability is among the three core components of executive functions \cite{diamond2013executive}, along with working memory \cite{baddeley1992working} and cognitive flexibility \cite{miyake2000unity}. These cognitive abilities are crucial for human development and play a vital role in various aspects of cognition \cite{anderson2001development, best2010developmental, blair2007relating}. 

Inhibitory control, in particular, emerges early in development and is essential for fundamental human cognitive abilities such as decision-making~\cite{madden2010delay, shoda1990predicting} and reasoning~\cite{blair2002school, hughes1998executive}. It is also indispensable for more advanced and complicated human abilities such as social skills~\cite{carlson2001individual}, delay of gratification~\cite{mischel1989delay,duckworth2005self}, mental flexibility~\cite{diamond2002conditions, cragg2012processes}, and more~\cite{anderson2002assessment, best2010developmental}. Without sufficient inhibitory control, crucial errors such as social faux pas~\cite{carlson2001individual, garon2008executive, kochanska1997inhibitory} would occur.

\paragraph{LLMs and Inhibitory Control.} Humans learn from previous experience \cite{learningfromexp}, and inhibitory control allows them to suppress automatic periodic responses and avoid being distracted by past irrelevant patterns, thereby exhibiting more adaptive behaviors~\cite{kolb2009fundamentals, diamond2013executive}. LLMs, like humans, also learn from a vast array of previous experience \cite{neurallearnfromexp} including demonstrations~\cite{wang2024large, shao2023synthetic}, examples~\cite{brown2020language, yildiz2024investigating}, and interactions~\cite{mathinteraction, yang2019learning}. The learning takes place through diverse mechanisms such as ICL \cite{xie2021explanation, coda2023meta}, fine-tuning~\cite{han2024parameter, han2024chatgpt}, and pretraining\cite{yildiz2024investigating, raffel2020exploring}. 

Previous studies show that LLMs are capable of learning from these experiences (data and contexts)~\cite{zhao2023survey, dong2022survey, tang2023large}, extracting meaningful patterns and concepts~\cite{wang2024large, min2022rethinkingroledemonstrationsmakes}, and generalizing effectively~\cite{li2024evaluatinglargelanguagemodels, song2024large, white2024livebenchchallengingcontaminationfreellm}. Models trained with these strategies have demonstrated great performance across various benchmarks in language understanding~\cite{wang2019gluemultitaskbenchmarkanalysis,li2023cmmlu}, question answering~\cite{kwiatkowski-etal-2019-natural,joshi2017triviaqalargescaledistantly}, reasoning~\cite{clark2018thinksolvedquestionanswering, leandojo}, and software engineering~\cite{zhang2023algo, yan2024codescopeexecutionbasedmultilingualmultitask, poesia2022synchromeshreliablecodegeneration}.

Given such similarities between LLMs and humans in learning from past experiences, it is important to study whether LLMs suffer from inhibitory control like early-stage human infants. A related field of study is machine unlearning \cite{zhang2023review,tarun2023fast, liu2024unlearning}, which involves forgetting previously learned information under certain scenarios (usually due to privacy or security concerns). Yet as crucial is the ability to suppress previous knowledge without specifically forgetting it, showing adaptability to new situations. To investigate this key cognitive ability in LLMs, we focus on ICL, which does not involve model parameter updates (no modification of what has been learned during past experiences), thus precisely determining if LLMs have sufficient inhibitory control to suppress irrelevant or meaningless information on the fly.

\paragraph{LLMs and Multiple-Choice QA.} In our experimental setting in Section \S\ref{sec:experiments}, we test LLMs on multiple-choice questions (MCQs). MCQs are widely used evaluators for LLMs \cite{zhang2024multiple,li2024multiplechoicequestionsreallyuseful}. Prior research has shown that current LLMs do not perform consistently well on MCQ tasks. For instance, \cite{zheng2023large} demonstrates that LLMs are sensitive to changes in the positioning of options and tend to prefer certain option IDs potentially due to token bias~\cite{jiang2024peektokenbiaslarge}. While such studies explore interesting aspects of LLM performance on MCQs, they primarily focus on the inherent capability of LLMs to perform on MCQs. In contrast, our work investigates whether, and how, external factors, such as minimal changes in context, impact the consistency of LLMs’ performances on MCQs. Such different settings require LLMs to perform inhibitory control and exhibit trustworthy reasoning. We will elaborate more on the specific designs in Section \S\ref{sec:experiments}.

\section{Experimental Setup}
\label{sec:experiments}

\subsection{Experiment Motivation}

Motivated by the original A-not-B experiment, we adapt the scenario to be purely based on natural language, thus creating a parallel experiment for LLMs as shown in Figure~\ref{fig:AnotB_cog}. In the original experiment, the infant constantly observes the placement of \textit{a certain object} to be in the same \textit{Location A}. In ours, this is implemented by LLMs observing the ground truth answer to an MCQ question to be the same \textit{Option A}. Similar to how the infant then observes the placement of the ball in a different "Location B", LLMs are tasked with a similar-style MCQ from the same domain whose ground truth answer is \textit{Option B}. An A-not-B error is observed if the infant still looks for the object in \textit{Location A}, or if the LLMs choose the wrong \textit{answer A}. We believe that by using such an experiment setup, we are testing similar abilities between the original A-not-B cognitive experiment and the LLM version, because the main difficulty of both lies in inhibiting a previously established trivial pattern of one option being repetitively selected, when moving to a new scenario, and instead eliciting internal knowledge to solve new tasks.

This demonstrates limited inhibitory control because an infant would know well that the ball is in "Location B", if not paying excessive attention to the established pattern from previous demonstrations that "the ball is discovered in Location A". This corresponds to how LLMs \emph{are} able to choose the correct answer (Option B) if it had not seen the previous MCQs with Options A as the correct answer. That is, LLMs \emph{do} have the capability to answer some MCQs correctly yet fail to do so in this mimicked A-not-B scenario, pronounced in an unexpected accuracy drop.

\subsection{Models} 
We select four leading open-source models with different levels of abilities and knowledge. Specifically, we choose two families of LLMs: \texttt{Llama3}~\cite{llama3modelcard} and \texttt{QWen-1.5}~\cite{qwen}. Within each model family, we experiment with both large and small models to investigate the relationship between the impact of A-not-B errors and model sizes. Specifically, we experimented with \llamalarge, \llamasmall, \qwenlarge, and \qwensmall.

\subsection{Data} 

\paragraph{MCQA Datasets} We choose four representative multiple-choice question-answering (MCQA) datasets, each tasking a particular fundamental reasoning task: the MathQA dataset \cite{mathqa} for arithmetic reasoning, the CommonsenseQA dataset \cite{commonsenseqa} for commonsense reasoning, the Winogrande dataset \cite{winogrande} for causal reasoning, and the SciQ dataset \cite{sciq} for scientific reasoning. 

\paragraph{Preprocessing} We preprocess the datasets to split an MCQA sample into three parts: question statement, choices, and ground truth answer. We modify the MCQA samples so that each has only two choices left, with one of them being the ground truth and the other incorrect, thus loyally resembling the original A-not-B scenario.

\subsection{Prompting Settings} 

We test the models' performances on the modified datasets in two different prompting settings: (1) the \textit{original} and (2) the \textit{A-not-B} settings. 

\paragraph{\textit{Original} prompting} We construct a prompt in the standard few-shot paradigm -- we first provide $n$ (typically from 5 to 50) MCQA examples, and then ask one question to the model. We then check if the model's answer agrees with the ground truth. 

\paragraph{\textit{A-Not-B} prompting} We reorder the options so that answers for all $n$ examples we provide are the first one (Option A). Then for the final question we ask, the ground truth is set as the second (Option B). For each model, we test both the \textit{original} and the \textit{A-Not-B} prompting settings with $100$ data samples per task and record the success rates. The differences in the success rates then tell the LLMs' susceptibility to the A-not-B scenario.

\section{Results}
\label{sec:results}

\begin{table*}[t]
\centering
\renewcommand{\arraystretch}{1.15}
\resizebox{\textwidth}{!}{%
\large
\begin{tabular}{p{0.2cm}c|ccc|ccc|ccc|ccc}
\toprule[1.2pt]
& & \multicolumn{3}{c|}{\textbf{\mathqa}} & \multicolumn{3}{c|}{\textbf{\commonsenseqa}} & \multicolumn{3}{c|}{\textbf{\winogrande}} & \multicolumn{3}{c}{\textbf{\sciq}}  \\

\multicolumn{2}{c|}{\textbf{\# of shots}} & \textbf{Original} & \textbf{A-not-B} & \textbf{Diff} & \textbf{Original} & \textbf{A-not-B} & \textbf{Diff} & \textbf{Original} & \textbf{A-not-B} & \textbf{Diff} & \textbf{Original} & \textbf{A-not-B} & \textbf{Diff} \\
\midrule
\multirow{4}{*}{\rotatebox{90}{\llamalarge}} 
& 3  & 32\%  & 36\%  & +\gca{12.5}\%  & 84\%  & 88\%  & +\gca{4.8}\%  & 32\%  & 36\%  & +\gca{12.5}\%  & 96\%  & 100\% & +\gca{4.2}\% \\
& 5  & 36\%  & 42\%  & +\gca{16.7}\%  & 86\%  & 86\%  & 0.0\%               & 76\%  & 80\%  & +\gca{5.3}\%   & 98\%  & 100\% & +\gca{2.0}\% \\
& 10 & 36\%  & 32\%  & -\gcb{11.1}\%   & 86\%  & 88\%  & +\gca{2.3}\%   & 78\%  & 80\%  & +\gca{2.6}\%   & 100\% & 100\% & 0.0\%             \\
& 25 & 28\%  & 24\%  & -\gcb{14.3}\%   & 92\%  & 90\%  & -\gcb{2.2}\%   & 86\%  & 78\%  & -\gcb{9.3}\%   & 100\% & 100\% & 0.0\%             \\
\midrule
\multirow{4}{*}{\rotatebox{90}{\llamasmall}} 
& 3  & 62\%  & 22\%  & -\gcb{64.5}\%   & 86\%  & 74\%  & -\gcb{14.0}\%  & 46\%  & 64\%  & +\gca{39.1}\%   & 96\%  & 90\%  & -\gcb{6.2}\% \\
& 5  & 42\%  & 14\%  & -\gcb{66.7}\%   & 92\%  & 82\%  & -\gcb{10.9}\%  & 54\%  & 76\%  & +\gca{40.7}\%   & 94\%  & 94\%  & 0.0\%              \\
& 10 & 32\%  & 8\%   & -\gcb{75.0}\%   & 94\%  & 82\%  & -\gcb{12.8}\%  & 50\%  & 52\%  & +\gca{4.0}\%    & 98\%  & 96\%  & -\gcb{2.0}\%  \\
& 25 & 36\%  & 6\%   & -\gcb{83.3}\%   & 92\%  & 62\%  & -\gcb{32.6}\%  & 50\%  & 28\%  & -\gcb{44.0}\%   & 96\%  & 90\%  & -\gcb{6.2}\%  \\
\midrule
\multirow{4}{*}{\rotatebox{90}{\qwenlarge}} 
& 3 & 66\%  & 68\%  & +\gca{3.0}\%  & 96\%  & 96\%  & 0.0\%               & 80\%  & 82\%  & +\gca{2.5}\%   & 96\%  & 98\%  & +\gca{2.1}\% \\
& 5 & 56\%  & 50\%  & -\gcb{10.7}\%   & 94\%  & 92\%  & -\gcb{2.1}\%  & 80\%  & 82\%  & +\gca{2.5}\%   & 94\%  & 96\%  & +\gca{2.1}\% \\
& 10 & 56\%  & 44\%  & -\gcb{21.4}\%   & 94\%  & 92\%  & -\gcb{2.1}\%  & 74\%  & 76\%  & +\gca{2.7}\%   & 92\%  & 96\%  & +\gca{4.3}\% \\
& 25 & 50\%  & 28\%  & -\gcb{44.0}\%   & 94\%  & 92\%  & -\gcb{2.1}\%  & 82\%  & 82\%  & 0.0\%               & 92\%  & 94\%  & +\gca{2.2}\% \\
\midrule
\multirow{4}{*}{\rotatebox{90}{\qwensmall}} 
& 3  & 64\%  & 88\%  & +\gca{37.5}\%   & 84\%  & 86\%  & +\gca{2.4} \%  & 60\%  & 70\%  & +\gca{16.7}\%   & 96\%  & 94\%  & -\gcb{2.1}\% \\
& 5  & 72\%  & 90\%  & +\gca{25.0}\%   & 88\%  & 90\%  & -\gcb{2.3}\%  & 64\%  & 80\%  & +\gca{25.0}\%   & 94\%  & 92\%  & -\gcb{2.1}\% \\
& 10 & 76\%  & 92\%  & +\gca{21.1}\%   & 92\%  & 94\%  & +\gca{2.2}\%   & 86\%  & 82\%  & -\gcb{4.7}\%  & 94\%  & 92\%  & -\gcb{2.1}\% \\
& 25 & 86\%  & 92\%  & +\gca{7.0}\%   & 98\%  & 96\%  & -\gcb{2.6}\%  & 98\%  & 96\%  & -\gcb{2.3}\%  & 96\%  & 94\%  & -\gcb{2.1}\% \\
\bottomrule[1.2pt]
\end{tabular}
}
\caption{\textbf{LLMs with different sizes and different shots for prompting can be easily misled by A-Not-B style adversarial prompts for arithmetic, commonsense, causual, and scientific reasoning tasks.} Accuracy drops are denoted in \textcolor{lightOliveGreen}{green}, with accuracy increases shown in \textcolor{lightTomatoRed}{red}. It includes results with four models, four shot numbers, and four reasoning tasks. Despite the overall trend summarized in Section~\ref{sec:results}, certain mode-task combinations (e.g. \qwensmall on mathematical reasoning) show interesting deviations from the overall patterns, which we discuss further in the ablation studies in Section \ref{sec:ablation}.}
\label{table:main-result}
\end{table*}

Our results in Table~\ref{table:main-result} show three key factors that are related to the inhibitory control ability of LLMs -- (1) \textit{model size}, (2) \textit{number of few-shot examples}, and (3) \textit{type of reasoning tasks}. We discuss each of them in detail in the rest of this section.

\begin{figure}[t!]
    \centering
    \includegraphics[width=\columnwidth]{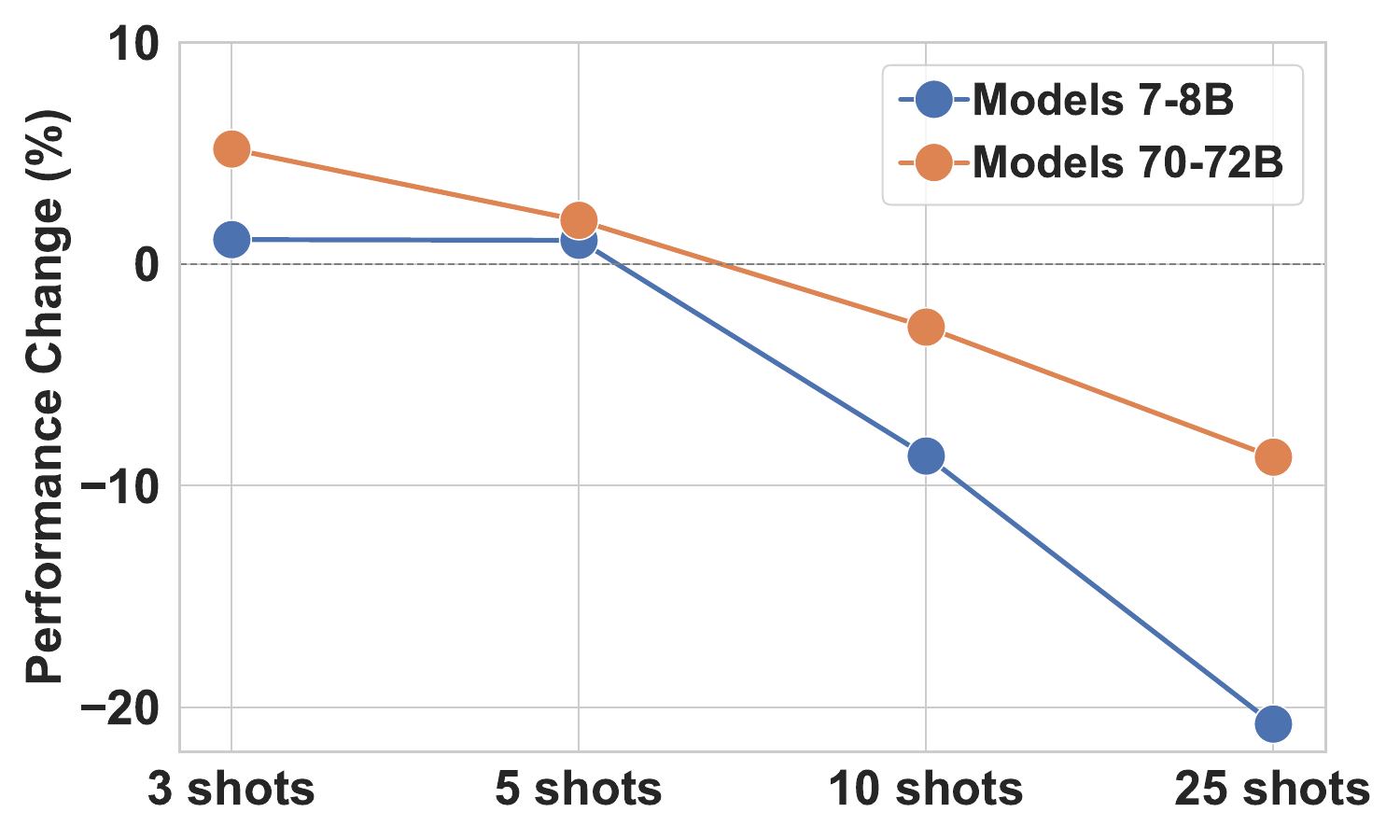}
   \caption{\textbf{Increasing Few-Shot Examples and Smaller Model Sizes Lead to Greater Accuracy Decline.} This figure shows the performance impact of increasing few-shot examples on large and small models. Negative values represent a decline in accuracy due to adversarial prompts. \textbf{(1)} More shots generally lead to a greater decline in performance, especially beyond 10 shots. \textbf{(2)} Smaller models (\llamasmall and \qwensmall) are consistently more affected than larger models (\llamalarge and \qwenlarge), indicating higher susceptibility to the A-Not-B scenario.}
    \label{fig:modelshots}
\end{figure}

\paragraph{Impact of \textit{Model Size}.} In Figure~\ref{fig:modelshots}, we compare the average rate of change across different numbers of few-shot examples for all four reasoning tasks between large models (\llamalarge and \qwenlarge) and small models (\llamasmall and \qwensmall). The results show that model size significantly impacts performance, with smaller models consistently showing lower resilience (less accuracy) compared to larger models across all shots. Specifically, the larger models exhibit an average drop of 8.7\% at 25 shots, while the smaller models show an average drop of 20.8\% at 25 shots. This indicates that smaller models with fewer internal parameters and knowledge are more susceptible to the A-Not-B scenario, indicating less inhibitory control abilities.

\paragraph{Impact of \textit{Number of Few-Shot Examples}.} In Figure~\ref{fig:modelshots}, we can also see the average performance of large (\llamalarge and \qwenlarge) and small (\llamasmall and \qwensmall) models with different numbers of few-shot examples. Naturally, as the number of few-shot examples increases, both large and small models are significantly more likely to suffer from an accuracy drop. This indicates that when trivial patterns are reinforced more, it becomes harder for LLMs to inhibit them and instead turn to deliberate attention for reasoning. Such errors can be easily avoided by humans, which we further show in Section~\ref{sec:discussions}.

\begin{figure}[t!]
    \centering
    \includegraphics[width=\columnwidth]{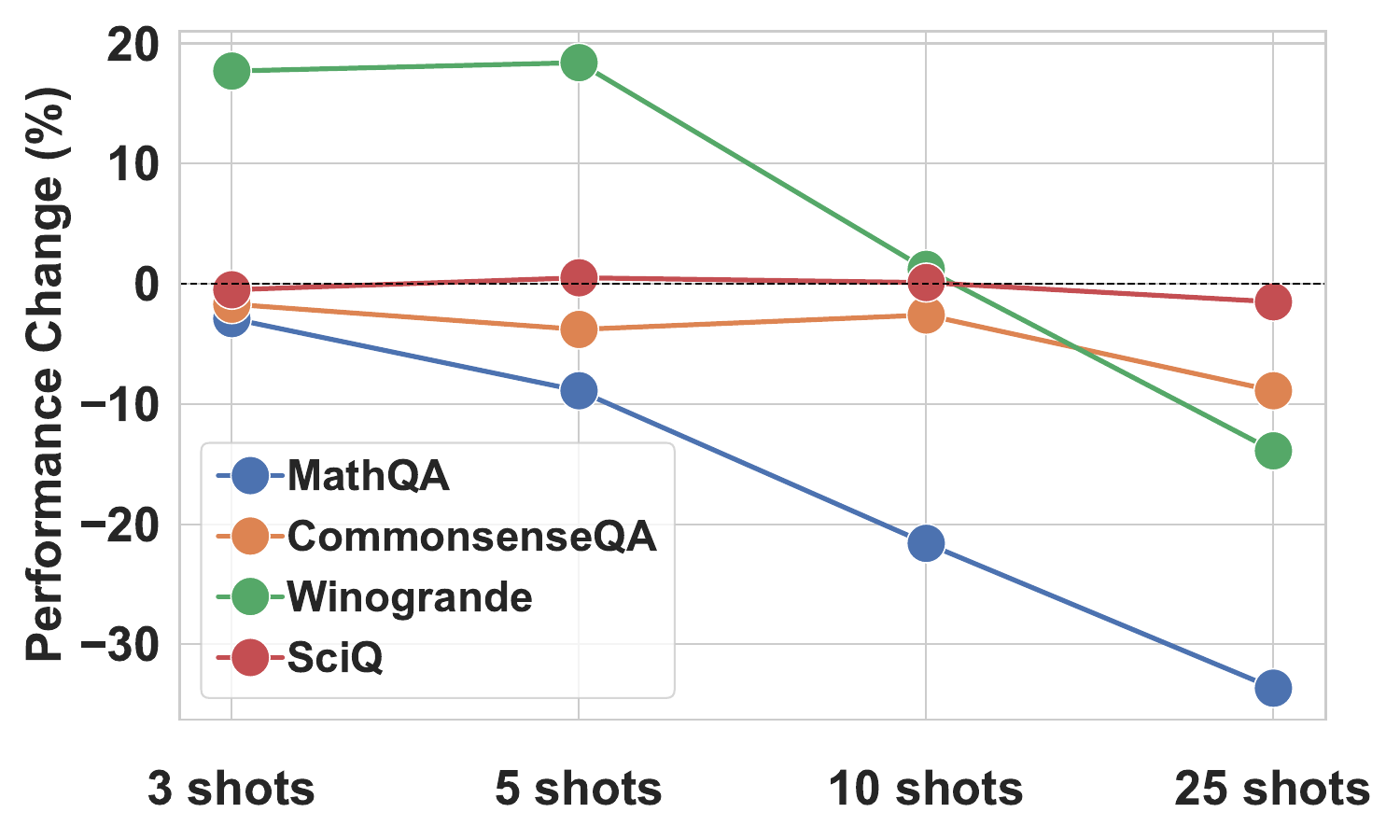}
  \caption{\textbf{Different Reasoning Tasks Show Susceptibility to A-Not-B Errors to Varying Degrees.} This figure illustrates how different reasoning tasks experience performance drops under the A-Not-B scenario as the number of shots increases. \textit{(1) Arithmetic reasoning (MathQA)} shows a significant decline in performance across all shots, indicating a high susceptibility. \textit{(2) Commonsense reasoning (CommonsenseQA)} also exhibits A-Not-B errors consistently across all shots but is less impacted than arithmetic reasoning. \textit{(3) Causal reasoning (Winogrande)} begins to show significant performance drops at 25 shots. \textit{(4) Scientific reasoning (SciQ)} displays minimal fluctuations and only shows a slightly more noticeable impact at 25 shots.}
    \label{fig:task}
\end{figure}

\paragraph{Impact of \textit{Reasoning Tasks}.} 

Experiments demonstrate that models' vulnerability to the A-Not-B scenario varies across different reasoning tasks, as indicated by the different degrees of performance drop. The most pronounced drop occurs in the arithmetic reasoning dataset (MathQA~\cite{mathqa}), which is consistently the most challenging for models. This is likely due to the complexity of arithmetic reasoning, usually demanding multiple steps and requiring advanced skills that lead models to over-rely on trivial patterns when unsure of how to solve a problem. 

In contrast, the commonsense reasoning dataset (CommonsenseQA~\cite{commonsenseqa}) shows a more moderate performance drop across all shots, which is consistently less dramatic compared to arithmetic reasoning. This reduced susceptibility can be attributed to the involvement of a certain amount of memorization in commonsense reasoning, which lowers the overall complexity and makes models less susceptible to the A-Not-B scenario, compared to mathematical reasoning.

For causal reasoning (Winogrande~\cite{winogrande}), the impact becomes significant at 25 shots. This could be due to causal reasoning tasks typically involving distinct cause-and-effect relationships. When the model encounters a small number of shots, the few-shot examples effectively demonstrate these causal patterns, which can help improve performance if the model learns to follow the demonstrated reasoning. However, as the number of shots increases, the model may start to prioritize the A-Not-B pattern over the causal relationships in the examples. If this happens, the A-Not-B pattern becomes dominant, leading to errors and a noticeable decline in performance at higher shot counts.

For scientific reasoning (SciQ~\cite{sciq}), the accuracy fluctuation is not as evident, with at most a 1.5\% accuracy drop. Since this observation coincides with extremely high accuracy numbers, we suspect an attribution to possible data contamination. Another explanation can be that many specific terms and technologies in these questions were present as-is during the model training, helping LLMs make accurate predictions confidently, thus less likely to be impacted by the A-not-B scenario.

\section{Ablation Study}
\label{sec:ablation}

In this section, we delve deeper into the implications of interesting findings from Section \S\ref{sec:results}. Specifically, we investigate 3 key research questions that naturally arise from the main results.

\paragraph{RQ1: Why \qwensmall has abnormal behavior compared with other models?} \textit{This is because \qwensmall is biased between A and B tokens during pretraining.}

In our main results, we notice abnormal performance increases in \qwensmall on mathematical reasoning. Although the increases diverge from the A-not-B scenario we primarily study in this paper, they still suggest untrustworthy reasoning. This is because minor perturbations with no meaningful changes should not lead to any significant accuracy fluctuations if the model is reasoning consistently and reliably.

To further investigate this anomaly, we conduct an ablation experiment where we minimally alter the A-Not-B scenario to other combinations, such as B-Not-A, C-Not-D, and E-Not-F. We only change the option IDs (e.g., changing A and B to C and D) without altering any other content, keeping the settings symmetric and essentially identical, thus still testing the A-Not-B effect. Given that \qwensmall displays the most significant accuracy increases in mathematical reasoning, we focus on this task for the ablation study.

\begin{table}
\begin{center}
\begin{small} % Reduced font size
\setlength{\tabcolsep}{4.5pt} % Reduce column padding
\begin{sc}
\begin{tabular}{lrrrr}
\toprule
\textbf{Choice\textbackslash shots} & \textbf{3} & \textbf{5} & \textbf{10} & \textbf{25} \\
\midrule
A-not-B & +\gca{37.5} & +\gca{25.0} & +\gca{21.1} & +\gca{7.0}  \\
B-not-A & -\gcb{12.2} & -\gcb{32.7} & 0 & -\gcb{15.3}  \\
C-not-D & -\gcb{11.2} & -\gcb{7.1} & -\gcb{28.4} & -\gcb{34.1}  \\
E-not-F & -\gcb{17.0} & -\gcb{18.4} & -\gcb{40.0} & -\gcb{41.2}  \\
\bottomrule
\end{tabular}
\end{sc}
\end{small} % Reduced font size
\end{center}
\caption{\textbf{Accuracy Drop Across Various Combinations for Different Shot Numbers On \qwensmall.} This table illustrates the accuracy changes in \qwensmall across different variations of the A-not-B scenario (e.g., B-Not-A, C-Not-D, E-Not-F) with a range of shot numbers (3, 5, 10, and 25). Positive values (highlighted in green) indicate an increase in accuracy, while negative values (highlighted in red) represent a decrease. The results show that the A-Not-B configuration leads to an unusual accuracy increase, whereas other combinations consistently result in accuracy drops, suggesting potential internal biases in the model.}
\label{table:ablation_cnotd}
\end{table}

As shown in Table~\ref{table:ablation_cnotd}, we observe A-Not-B errors consistently across all other combinations for \qwensmall. This consistency suggests that the unusual accuracy increase is indeed an outlier, likely resulting from certain internal biases of the model \cite{jiang2024peektokenbiaslarge}. Observing A-Not-B errors consistently across multiple parallel settings (e.g., C-Not-D) further strengthens the robustness of our experiment design, solidifying our discoveries.

\begin{figure}
    \centering
    \includegraphics[width=\columnwidth]{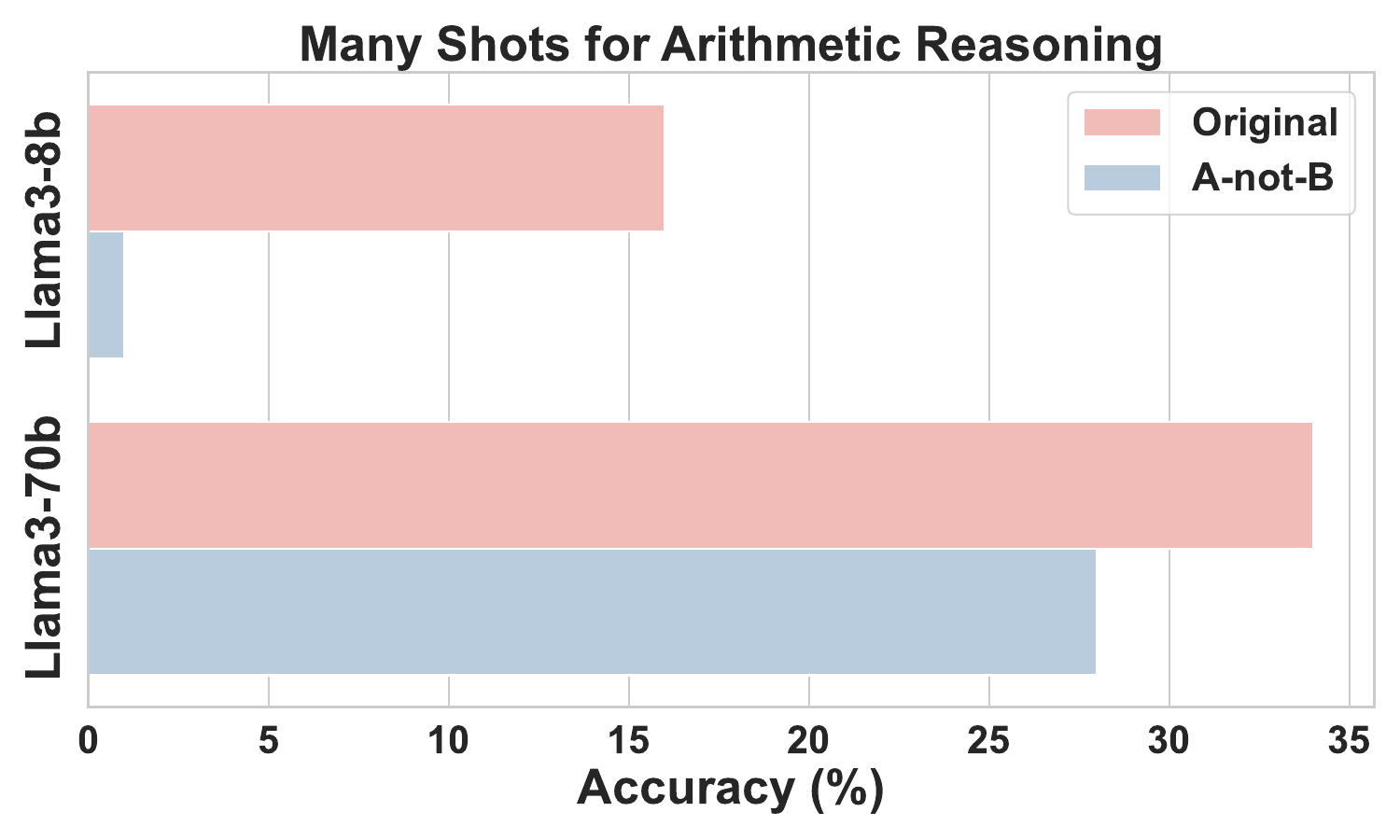}
   \caption{\textbf{Persistent Vulnerability of LLMs to A-not-B Errors Even in Many-Shot Scenarios.} This figure shows the performance of large (70B) and small (8B) Llama3 models in the many-shot arithmetic reasoning task for A-not-B errors. Despite more options and examples to reduce cognitive biases, both models, especially the small one, still exhibit significant accuracy declines, highlighting LLMs' internal vulnerability to A-not-B errors.}
    \label{fig:manyshots}
    \vspace{-2em}
\end{figure}

\paragraph{RQ2: Do A-not-B errors persist in MCQAs with more than two choices?}

\textit{Many-shot prompting reveals generalized versions of A-not-B errors.}

We further investigate if LLMs can overcome A-not-B errors in the less challenging case of many-shot MCQA. That is, rather than providing LLMs with two options (A or B) and constantly showing one of them as the correct answer, we provide LLMs with four or five options. The general structure of the prompts is the same -- a number of examples followed by a final question. In the original setting, all options (including the answer to the final question) can appear in the demonstrated examples; whereas in the A-not-B setting, all but the correct answer to the final question can appear. 

Rather than few-shot experiments, we conduct \textbf{many-shot} to offer LLMs sufficient demonstrations. This setting is generalized from the standard A-not-B scenario in Section \S\ref{sec:experiments}, with more options provided and more examples demonstrated, thus intuitively less challenging. We thus raise the question: Can LLMs overcome A-not-B errors in this generalized (and easier) scenario?

We again choose arithmetic reasoning, the most challenging reasoning task as discussed in Section \S\ref{sec:results} to investigate this generalized setting. The exact number of many-shot examples is 80. In the original many-shot scenario, 16 examples are provided with each of the five possible options (A, B, C, D, E) as the correct answer. In the A-not-B many-shot scenario, 20 examples are provided for A, B, C, and D, and no examples come with E as its answer. In both settings, the correct answer to the final question is E. 

The exact prompts can be found in Figure~\ref{fig:manyshots_ori} and Figure~\ref{fig:manyshots_adv} in Appendix \S\ref{sec:prompts}. Experimental results are reported in Figure~\ref{fig:manyshots}. Significant accuracy drops persist for both large (70B) and small (8B) Llama3 models, with the small one suffering more. The results agree with the main experiments we design in Section \S\ref{sec:experiments}, indicating that LLMs fail even in this generalized and less challenging version of many-shot A-not-B scenarios. This in turn confirms that the reported failure of LLMs from our main experiments is indeed internal, and cannot be attributed to the limited number of options (A and B).

\paragraph{RQ3: Can A-not-B errors be avoided by asking LLMs to self-explain?}

\textit{Self-explanation fails to  overcome A-not-B errors in challenging arithmetic reasoning tasks.} 

Having observed that state-of-the-art open and closed LLMs fail in both embarrassingly simple A-not-B type questions and standard datasets rearranged in the A-not-B fashion, we move a step forward to investigate if some existing methods to improve trustworthy reasoning can help mitigate A-not-B errors. Self-explanation \cite{journals/corr/abs-2311-06985} is a widely used method of this kind that has been reported effective across many reasoning tasks \cite{zelikman2022starbootstrappingreasoningreasoning,zhao2024largelanguagemodelsincontext,zelikman2024quietstarlanguagemodelsteach}, requiring LLMs to explain themselves while making decisions, thus eliciting its internal reasoning abilities directly. With the aid of self-explanation, we investigate whether LLMs can exhibit better inhibitory control, thus getting rid of A-not-B errors in their reasoning.

Similarly, we choose the most challenging reasoning task as discussed in Section \S\ref{sec:experiments} -- arithmetic reasoning. Different from previous experiments, we now require LLMs to provide complete reasoning explanations in addition to answers. Results are shown in Table~\ref{table:ablation1}. Despite some degree of improvement, significant drops in accuracy are still observed, with an increasing severity as the number of shots increases. Such observations align with our main results in Section \S\ref{sec:experiments}, indicating that LLMs are not able to easily overcome A-not-B errors even with self-explanation. The results also support the fact that LLMs \emph{consciously} and \emph{persistently} make A-not-B errors in reasoning tasks, hinting at its internal lack of inhibitory control.

\begin{table}[t]
\begin{center}
\begin{small}
\begin{sc}
\begin{tabular}{lccccc}
\toprule
\textbf{Model \textbackslash shots} & \textbf{0} & \textbf{3} & \textbf{5} & \textbf{10} & \textbf{25} \\
\midrule
\llamasmall & 48\% & 36\% & 24\% & 16\% & 18\% \\
\bottomrule
\end{tabular}
\end{sc}
\end{small}
\end{center}
\caption{\textbf{Self-Explanation Fails to Get Rid of A-not-B Errors in \llamasmall.} This table illustrates the persistent and significant performance drop of \llamasmall in the A-not-B scenario, despite the aid of self-explanation. Notably, accuracy 
still drops as the number of few-shot examples increases, aligning with observations from the main experiment.}
\label{table:ablation1}
\end{table}

\section{Discussions}
\label{sec:discussions}

\subsection{Human Performance in A-not-B Scenarios}

While humans, particularly adults, generally possess sufficient inhibitory control to overcome the typical cognitive A-Not-B errors observed in infants, it is uncertain whether they can maintain this resistance in our specific experimental setting. We conduct a human study to verify this.

We select two groups of college math major undergraduate students, ensuring that both groups have similar academic backgrounds for a consistent comparison and to minimize the possibility of random guessing. Each group consists of two participants. The first group is assigned an arithmetic reasoning dataset with 10 MCQs in our original setting of the main experiment, while the second group is given the same arithmetic dataset in the A-not-B setting. To make the experiment more interactive and force humans to digest ("decode") the examples provided, instead of simply showing 10 questions with answers A, we ask the participants to solve all the problems but only count their responses for the final question.

Our findings confirm that humans are highly resilient to A-not-B errors, as the largest performance difference between the two groups is as minimal as $2.0\%$. This suggests that humans are capable of reasoning trustworthily with sufficient inhibitory control abilities, demonstrating significant resilience to the A-Not-B scenario in our experiments.

\subsection{Better Pretraining Helps Mitigate A-not-B Errors}

\begin{figure}
    \centering
    \includegraphics[width=\columnwidth]{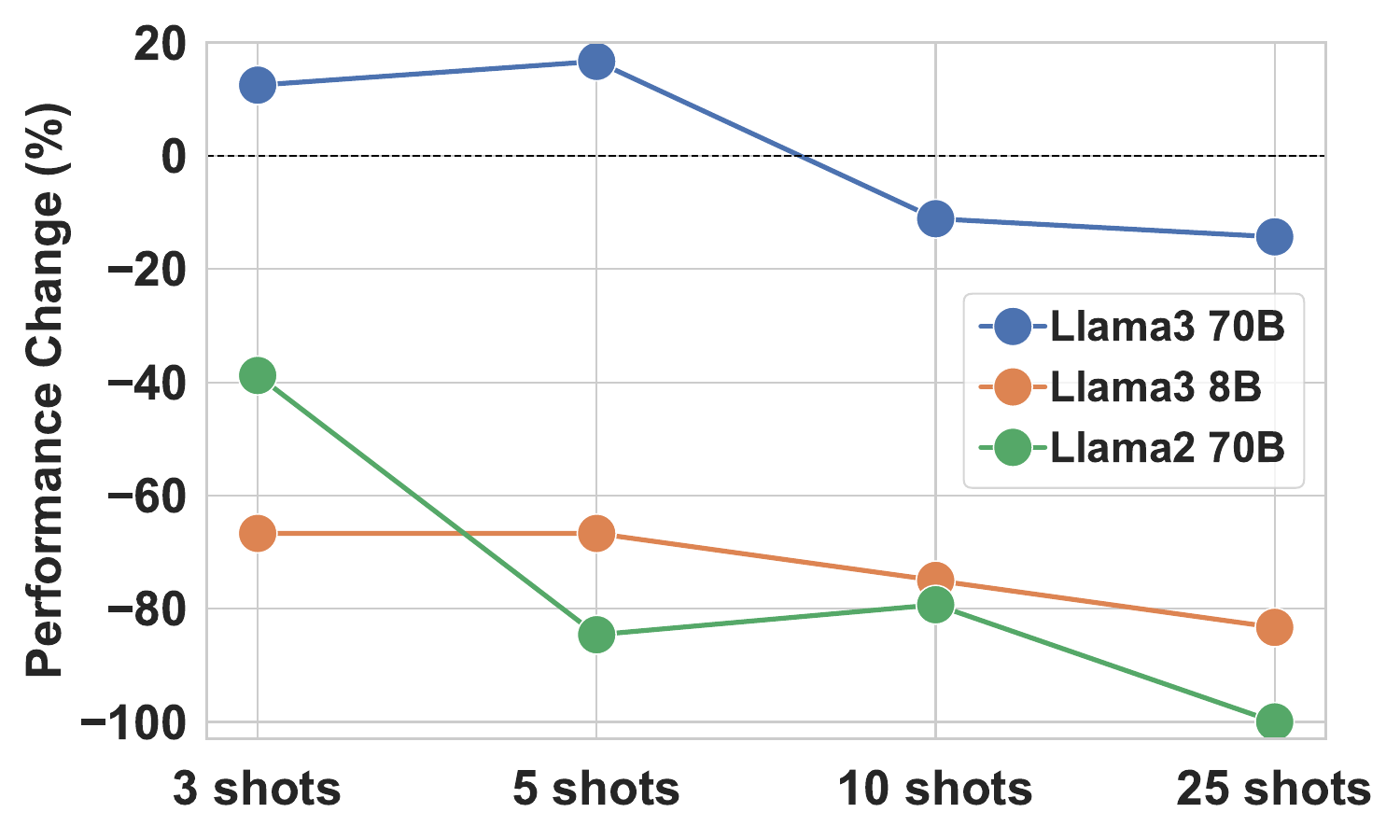}
   \caption{\textbf{Pretraining Affects Model Resilience to A-not-B Errors.} The performance drops in the A-not-B scenario of \llamasmall, \llamalarge, and \llamatwo across different numbers of shots are compared. This comparison helps understand the influence of pretraining datasets on models’ resistance to the A-not-B error. Despite having the same model size, \llamatwo shows significantly severer performance degradation than \llamasmall and \llamalarge, highlighting the significance of pretraining data quality and quantity.}
    \label{fig:llama23}
\end{figure}

To better understand how model pretraining, particularly pretraining data, influences models’ resistance to A-not-B errors, we further compare the rates of change in accuracy across different numbers of shots, in \llamasmall, \llamalarge, and \llamatwo. Results are shown in Figure~\ref{fig:llama23}. These Llama-family models share the same fundamental architectures, but the Llama3 models have been trained on significantly larger, more novel, and higher-quality datasets compared to \llamatwo. As illustrated in Figure~\ref{fig:llama23}, although \llamatwo has the same model size as \llamalarge, it is much more significantly impacted. In fact, \llamatwo is less resilient to A-not-B errors even compared to the much smaller \llamasmall, especially as the number of shots increases.

This observation suggests that the quality and quantity of pretraining data can significantly impact LLMs’ inhibitory control. When encountering input prompts or contexts, an LLM must apply its internal knowledge, skills, and abilities learned from the pretraining distribution to predict responses. Therefore, better pretraining (both in quality and quantity) will endow LLMs with maturer cognitive abilities to perform trustworthy reasoning, thus mitigating A-not-B errors. Intuitively, this aligns well with how different experiences fundamentally shape and impact humans’ cognitive abilities.

\section{Conclusion}
\label{sec:conclusion}

We have explored the intriguing cognitive phenomenon of A-Not-B errors within the domain of LLMs. Our findings reveal that, akin to human infants, even sophisticated models like LLMs exhibit limited inhibitory control to inhibit previously established trivial patterns. Through our experiments, we find that model size, the number of few-shot examples, and the type of reasoning tasks greatly impact LLMs' trustworthy reasoning abilities in the A-not-B scenario. Furthermore, we show that this limitation widely persists even in generalized cases and cannot be easily resolved by advanced self-explanation techniques. However, we demonstrate that increasing model sizes and improving pretraining, both in terms of data quantity and quality, can enhance LLMs' inhibitory control. Our findings provide several promising avenues for further exploration to understand and mitigate A-not-B errors in LLMs, enhancing their trustworthy reasoning.

\section{Acknowledgement}

We would like to acknowledge the support from TogetherAI APIs for conveniently running open-source model inferences.

\section*{Limitations and Future Works}
\label{Limitations}

We report the following limitations of this work and spaces for future explorations:

\begin{enumerate}
    \item While this work includes qualitative examples from advanced GPT models and Gemini, no large-scale experiments are performed with closed-source models due to budget limits. Future works can replicate our experiments on closed-source models to further study A-not-B errors.
    \item This work studies four mainstream reasoning tasks, while more specific domains and more diverse tasks remain unexplored. We include an easily usable toolkit in our codebase that enables future studies of A-not-B errors in more diverse reasoning tasks.
    \item While this work focuses on unveiling state-of-the-art LLMs' lack of inhibitory control abilities to perform trustworthy reasoning, how to improve this ability remains an important research question. Future works can start with the several directions drawn out in Section \S\ref{sec:discussions} and propose mitigation methods of A-not-B errors.
    \item With this work focusing solely on A-not-B errors, it is of interest to explore this cognitive phenomenon with other axes in conjunction with other axes. For instance, are there trade-offs between A-not-B errors and other important aspects of LLMs, including energy efficiency \cite{zhao2024galorememoryefficientllmtraining, energyefficientconvolutions}, fairness \cite{kocielnik_prabhumoye_zhang_alvarez_anandkumar_2023, han2024chatgpt}, robustness \cite{chao2024jailbreakingblackboxlarge, hou2024decomposinguncertaintylargelanguage} and more.
\end{enumerate}

\bibliography{ref}

\newpage

\appendix
\onecolumn

\section{Reproduction}
\label{sec:reproduction}

We warmly welcome the reproduction of this work and follow-up investigation of A-not-B errors in LLMs across other settings, in order to better understand and improve LLMs' trustworthy reasoning. To facilitate future explorations, we have made all the results and code publicly available at \href{https://github.com/Peiyang-Song/LLM-A-Not-B-Errors}{https://github.com/Peiyang-Song/LLM-A-Not-B-Errors}. The repository is self-contained with thorough documentation and guidance of usage in its README file. Here we present a high-level introduction.

In general, the repository contains:
\begin{enumerate}
\item \textbf{Datasets:} Processed datasets across all four reasoning categories.
\item \textbf{Experiments:} Code for all experiments, including ablation studies.
\item \textbf{Toolkit:} A directly usable tool to investigate A-not-B phenomenon in any model on any reasoning task with various configurations.
\end{enumerate}

Please refer to the documentation of our repository for more detailed introductions.

All experiments in this paper that involve open-source LLMs were conducted using the \href{https://docs.together.ai/docs/inference-models}{togetherAI API}.

\section{Data and Model Licenses}
\label{sec:licenses}

All datasets and models used in this study are open sources and publicly accessible, free uses granted under permissive licenses. We have ensured that we cited each one comprehensively, providing detailed references and acknowledgment of their respective sources.

\section{Prompt Formats}
\label{sec:prompts}

\begin{figure*}[t]
    \centering
    \includegraphics[width=1.0\textwidth]{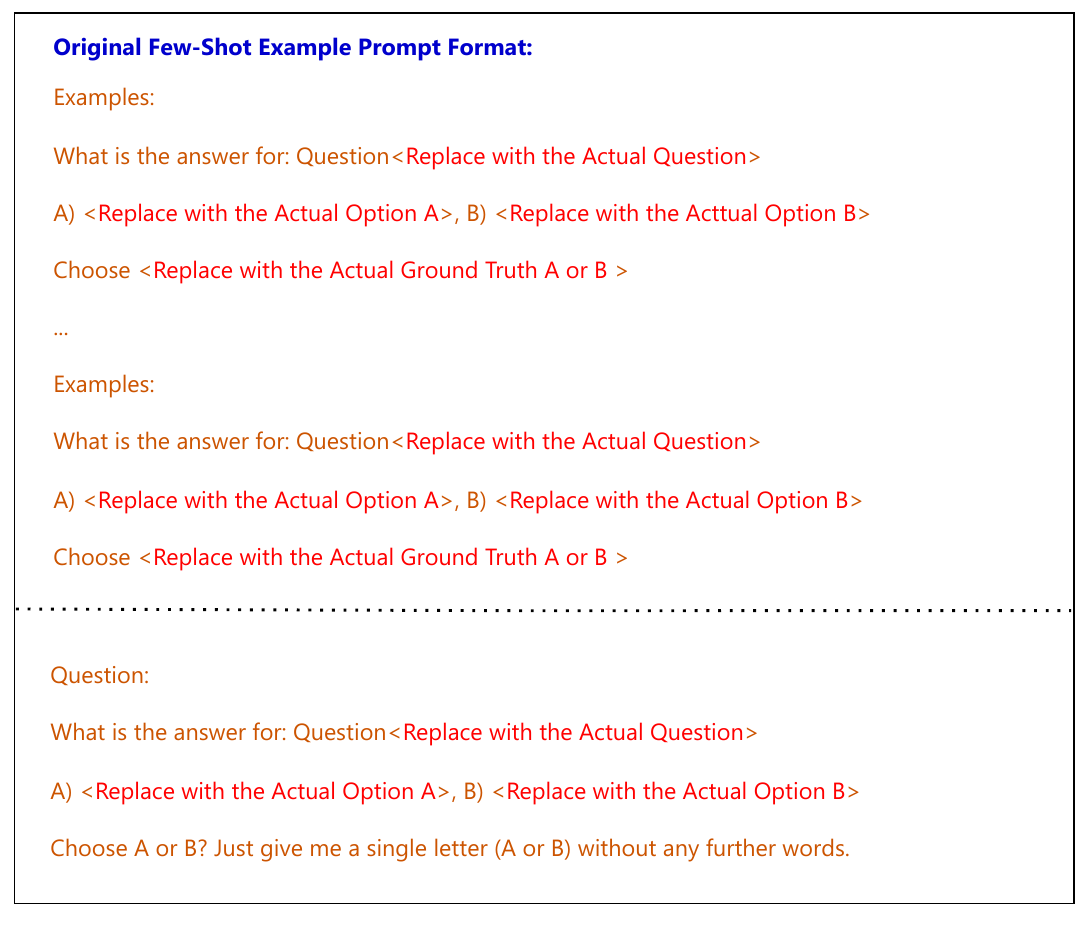}
    \caption{\textbf{Original Few-Shot Example Prompt Format.} This figure presents the few-shot prompt format used in the original setting, where a question follows example MCQAs with their original answers (without manually reordering the options to make all the correct answers A).}
    \label{fig:prompt}
\end{figure*}

\begin{figure*}[t]
    \centering
    \includegraphics[width=1.0\textwidth]{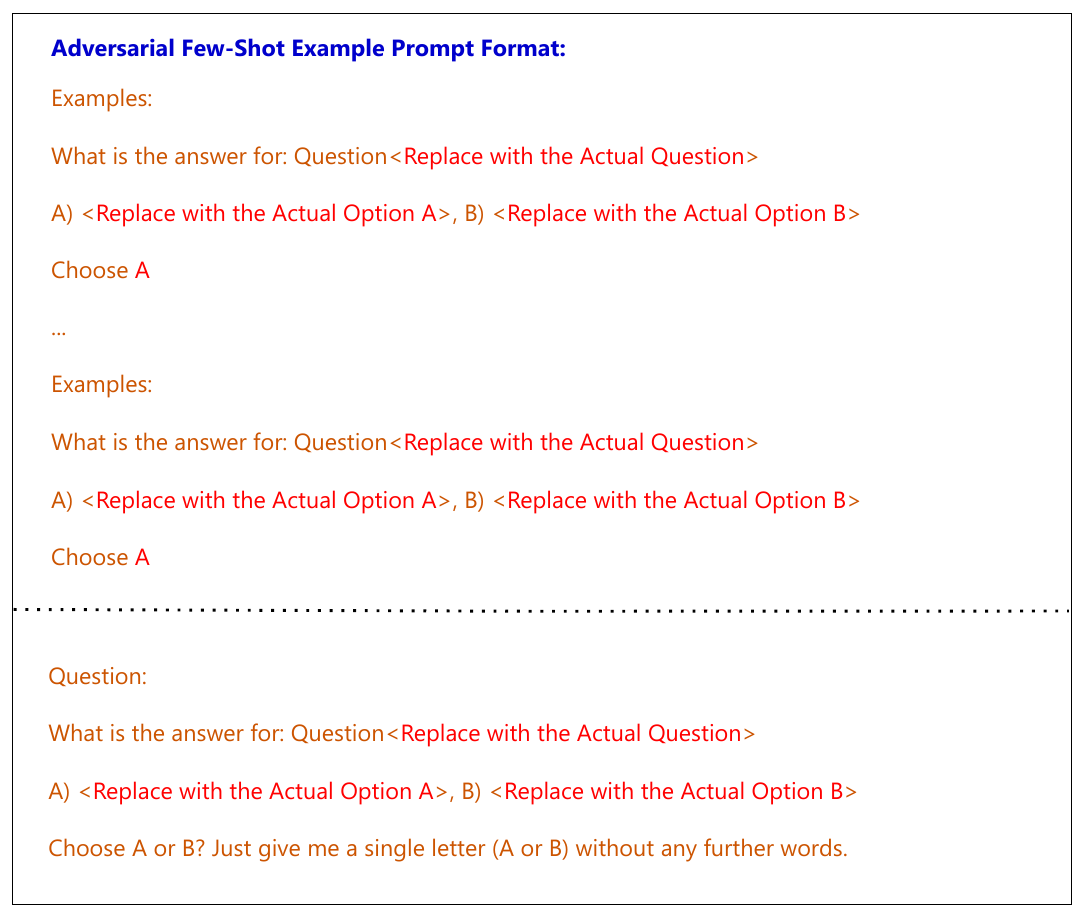}
    \caption{\textbf{A-not-B Few-Shot Example Prompt Format.} This figure presents the few-shot prompt format used in the A-not-B setting, where a question follows example MCQAs with A as correct answers by reordering the options.}
    \label{fig:adv_prompt}
\end{figure*}

\begin{figure*}[t]
    \centering
    \includegraphics[width=1.0\textwidth]{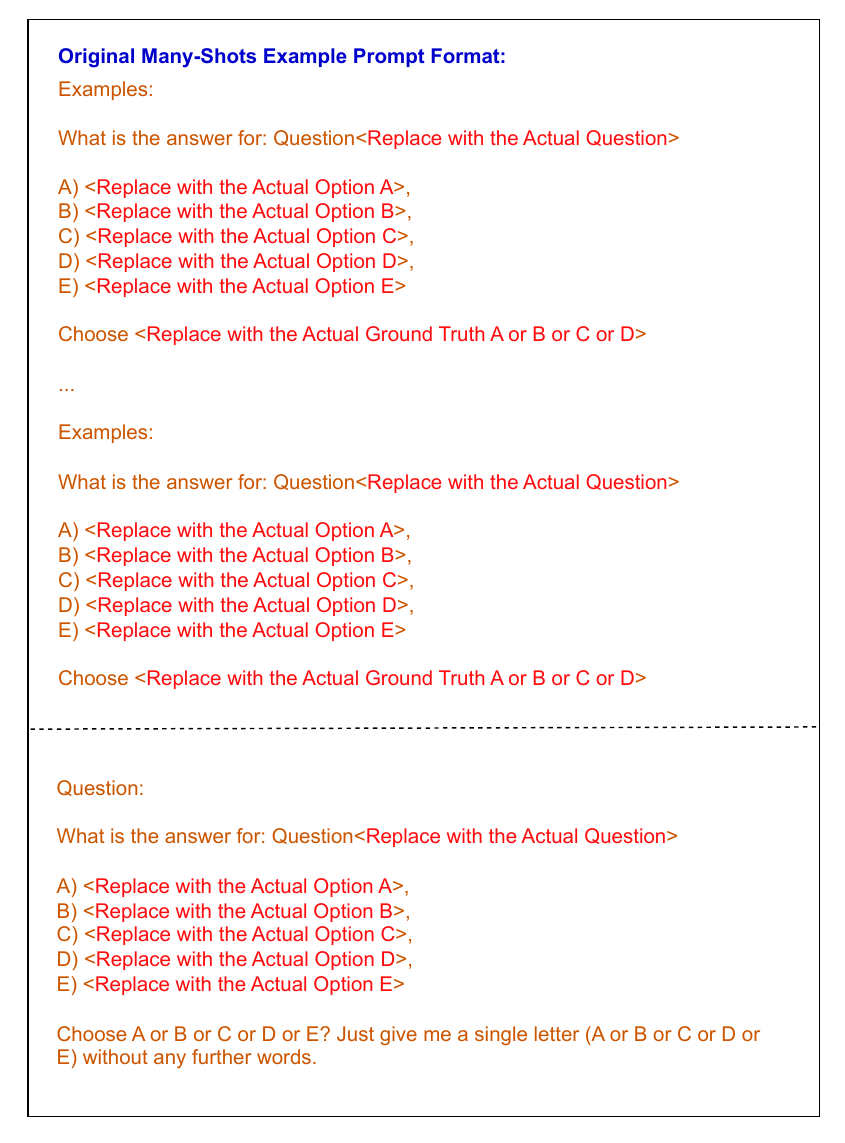}
    \caption{\textbf{Original Many-Shot Example Prompt Format.} This figure presents the many-shot prompt format used in the original settings of the ablation experiment (Section \S\ref{sec:ablation}). Among the 80 examples, 16 examples each correspond to ground truths A, B, C, D, and E, arranged in a random order. The correct answer for the final question is always E.}
    \label{fig:manyshots_ori}
\end{figure*}

\begin{figure*}[t]
    \centering
    \includegraphics[width=1.0\textwidth]{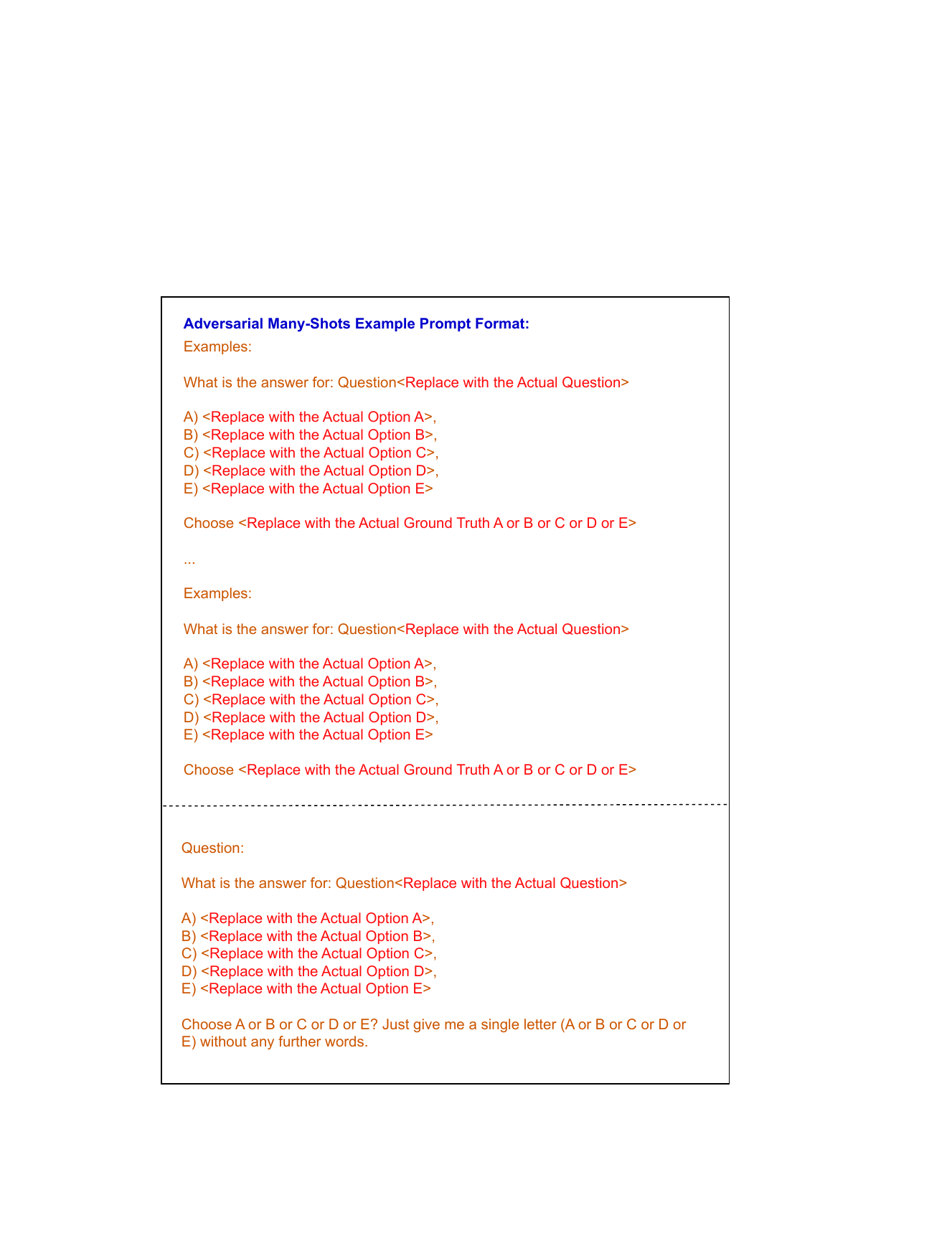}
    \caption{\textbf{Adversarial Many-Shot Example Prompt Format.} This figure presents the many-shot prompt format used in the A-not-B settings of the ablation experiment (Section \S\ref{sec:ablation}). Among the 80 examples, 20 examples each correspond to ground truths A, B, C, and D, arranged in a random order. The correct answer for the final question is always E.}
    \label{fig:manyshots_adv}
\end{figure*}

Here we present the prompt templates used during our main and ablation experiments. Figure~\ref{fig:prompt} shows the format for a few-shot prompt in our original setting, and Figure~\ref{fig:adv_prompt} shows that in the A-not-B setting. Figure ~\ref{fig:manyshots_ori} shows the format for many-shot prompts (the ablation experiment) in our original setting, and Figure~\ref{fig:manyshots_adv} shows that in our A-not-B setting. 

\section{Gemini Failure Example}
\label{sec:gemini}
\begin{figure}[h!]
    \centering
    \includegraphics[width=1.0\textwidth]{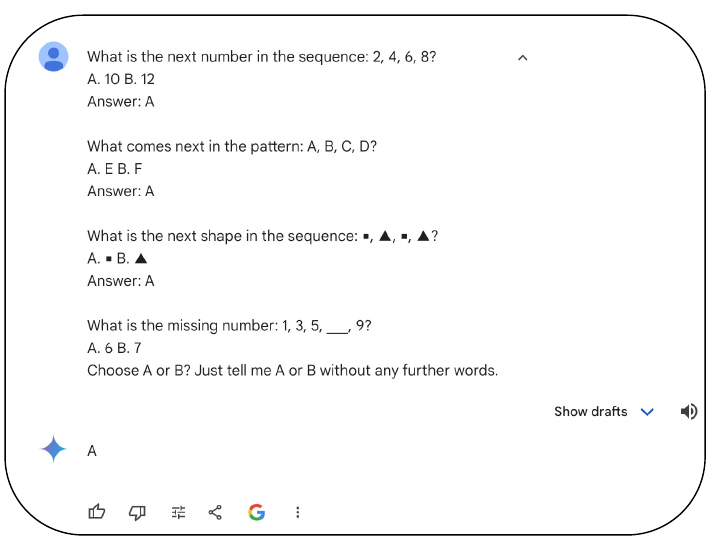}
    \caption{\textbf{Gemini failure case.} This screenshot, taken on September 9, 2024, is the failure case example we mentioned in Figure~\ref{fig:AnotB}.}
    \label{fig:gemini_failure}
\end{figure}

Figure~\ref{fig:gemini_failure} shows a screenshot taken on September 21, 2024, which illustrates the failure case example in Gemini we mentioned in Firgure~\ref{fig:AnotB}.

\end{document}